\documentclass[11pt, twocolumn]{article}

\usepackage{authblk}

\makeatletter
\renewcommand{\maketitle}{\bgroup\setlength{\parindent}{0pt}
\begin{flushleft}
  \textbf{\@title}

  \@author
\end{flushleft}\egroup
}
\makeatother

\title{\Huge Efficient Baseline for Quantitative Precipitation Forecasting in Weather4cast 2023  \vspace{0.5cm}}
\author{\Large Akshay Punjabi, Pablo Izquierdo Ayala \normalsize \hspace{0.5cm}}

\date{\today}

\usepackage[showframe=false, margin=.75in]{geometry}

\usepackage{abstract}
\renewcommand{\abstractname}{}    

\usepackage{lipsum}

\usepackage{afterpage}

\renewenvironment{abstract}
 {\small
  \begin{center}
  \bfseries \abstractname\vspace{-.5em}\vspace{0pt}
  \end{center}
  \list{}{%
    \setlength{\leftmargin}{10mm}
    \setlength{\rightmargin}{\leftmargin}%
  }%
  \item\relax}
 {\endlist}

\usepackage[utf8]{inputenc}

\usepackage{wrapfig}
\usepackage[lofdepth,lotdepth]{subfig}
\usepackage{booktabs}

\usepackage{rotating}

\usepackage{caption}

\usepackage{csquotes}

\usepackage{todonotes}

\usepackage{stfloats}

\usepackage[T1]{fontenc}
\usepackage{textcomp}
\usepackage{times}

\usepackage{framed} 

\usepackage[citestyle=ieee,
			bibstyle=ieee,
			language=auto,
			url=false,
			backend=bibtex,
			doi=false,
			isbn=false]{biblatex}
	\renewbibmacro*{volume+number+eid}{%
  \printfield{volume}%
  \setunit*{\addnbspace}
  \printfield{number}%
  \setunit{\addcomma\space}%
  \printfield{eid}}
\DeclareFieldFormat[article]{number}{\mkbibparens{#1}}
\usepackage{setspace}
\usepackage{color}
\definecolor{black}{gray}{0} 

\usepackage[colorlinks=true,linkcolor=black,citecolor=black]{hyperref}

\usepackage{tabularx}
\newcolumntype{b}{X}
\newcolumntype{s}{>{\hsize=.5\hsize}X}

\usepackage{ntheorem}

\begin{document}

\twocolumn[
\begin{@twocolumnfalse}
\maketitle
\begin{abstract}
\textit{Accurate precipitation forecasting is indispensable for informed decision-making across various industries. However, the computational demands of current models raise environmental concerns. We address the critical need for accurate precipitation forecasting while considering the environmental impact of computational resources and propose a minimalist U-Net architecture to be used as a baseline for future weather forecasting initiatives.}
\end{abstract}
\end{@twocolumnfalse}
]

\section{Introduction}
Accurate precipitation forecasting is critical for decision making across many industries, such as agriculture, transportation, energy generation, and more. At the same time, extreme weather events like storms, flooding, and drought can have devastating humanitarian and economic consequences if not predicted and prepared for effectively\cite{2022_agyekum_etal_ContributionWeatherForecast}.\newline

While physics-based numerical weather prediction (NWP) models have been the primary approach to forecasting, they require enormous computational resources. Recently, data driven deep learning methods have shown promise for weather forecasting\cite{2023_ben-bouallegue_etal_RiseDatadrivenWeather}, especially for short-term predictions. However, the computational cost of training these models scales with the number of parameters they contain, raising the concern about the derived environmental costs\cite{2019_strubell_etal_EnergyPolicyConsiderations}.\newline

In this paper, we present an efficient deep learning approach to this precipitation forecasting task using a variant of the U-Net architecture. This model provides a low resource and efficient alternative that can serve as a baseline for future weather forecasting initiatives.

\section{Weather4cast 2023}

The Weather4cast challenge\cite{2021_gruca_etal_CDCEO21First}\cite{2021_herruzo_etal_HighresolutionMultichannelWeather}\cite{2022_gruca_etal_Weather4castNeurIPS2022}, in its 2023 edition, was run by the Silesian University of Technology (Poland) in collaboration with the Johannes Kepler University (Austria) and the State Meteorological Agency in Spain (AEMET). This year's edition focuses on using multi-spectral satellite image data to forecast high-resolution, quantitative rain rates. The data provided was from 7 European regions selected based on their precipitation characteristics, ranging from February to December 2019 and January to December 2020. There are 3 different competition modalities:\begin{itemize}
    \item Core competition: This is the original Weather4cast competition. The task was to predict the exact amount of rain events 8 hours into the future from a 1 hour sequence of satellite images.
    \item Nowcasting competition: A reduced version of the original competition, with a target prediction of a 4 hours timeframe into the future.
    \item Transfer Learning competition: For this competition, test data for additional 3 unseen regions ranging from 2019 to 2020 was provided. The objective of this challenge was to make a 4 hour prediction on these regions, as well as a prediction on the 7 original regions for 2021.
\end{itemize}

We participated on the Nowcasting challenge, as we considered the 4 hour time frame an extremely practical forecast horizon for many applications, requiring the capturing of complex spatio-temporal interactions from relatively sparse and noisy input data. We decided not to participate in the 8 hour challenge, as based on our previous experiences in weather forecasting, the quality of the prediction degrades significantly past the first 3 hours and extending our prediction frame to 8 hours would only increase the complexity of our model and training times without providing any additional benefit.\newline

\section{Methods and Data}
\subsection{Model} \label{model}
Our model strucutre is based on a Small Attention-UNet (SmaAt-Unet)\cite{2021_trebing_etal_SmaAtUNetPrecipitationNowcasting}\cite {2021_punjabi_ayala_EfficientSpatiotemporalWeather}, which is an efficient variant of the widely-used U-Net architecture\cite{2015_ronneberger_etal_UNetConvolutionalNetworks}.\newline 

The key adaptations in SmaAt U-Net with respect to the original U-Net model are two:
\begin{itemize}
    \item Replacing all standard convolutional layers with depthwise separable convolutions (DSC). These factorize a convolution into a depthwise spatial convolution and a 1x1 pointwise convolution, reducing computations.
    \item Incorporating a channel attention mechanism (CBAM) to emphasize informative features. This sequentially applies channel and spatial attention to the feature maps.
\end{itemize}

Our implementation uses a baseline encoder-decoder U-Net architecture with skip connections. The encoder contains 5 stages with max pooling to downsample, while the decoder upsamples to recreate the input resolution. The input and output have 11 and 1 channels respectively corresponding to the satellite image bands and rain rate prediction.\newline

This model contains $\sim$4.1M parameters, in contrast to the $\sim$22M parameters of the provided baseline model.
We used AdamW optimizer with learning rate of 1e-3, batch size of 32 and early stopping monitoring validation loss. All models were trained using 2 NVIDIA T4 GPUs.

\begin{figure*}[ht!]
\centering
\includegraphics[width=1\textwidth]{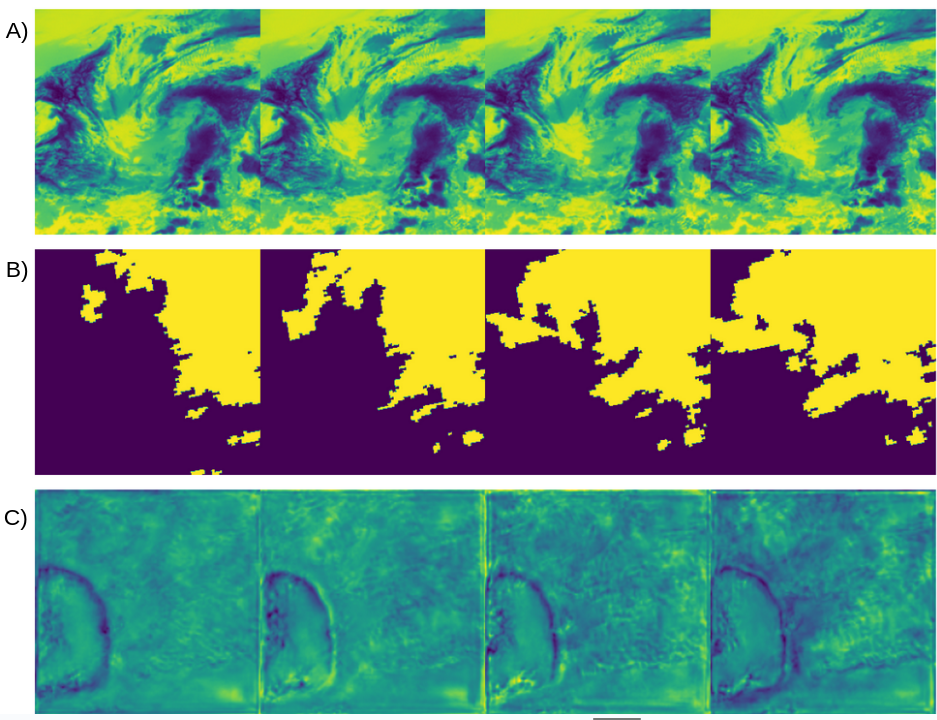}
\caption{Data example of the region roxi0007. A, input sequence corresponding to IR20 satellite band. B, the ground-radar reflectivity mask which corresponds to the prediction target. C, the predicted output of our model.}
\label{fig:data}
\end{figure*}

\subsection{Data}
We were provided with data from 7 different European regions. Four sets of 11-band spectral satellite images form the input sequence, with each band showcasing slightly noisy satellite radiances in visible (VIS), water vapor (WV), and infrared (IR) bands. Each satellite image spans a 15-minute period, and the pixels cover a spatial area of about 12km x 12km. The output prediction comprises a sequence of 16 images representing rain rates based on ground-radar reflectivities. These output images maintain a 15-minute temporal resolution but offer a heightened spatial resolution, with each pixel relating to an area of around 2km x 2km (See Fig.\ref{fig:data}).

\subsection{Data processing} 
To facilitate learning given the imbalanced rain distribution, we filtered, cropped and resized the raw input data:

\begin{itemize}
    \item Non-Rainy Filtering: Samples with total rainfall volumes below a threshold are considered effectively non-rainy and removed. This retains a fraction of the original training data but focuses learning on informative rain examples.
    \item Center Cropping: Given that the ground-radar rain rates region corresponds to a 42x42 pixel center region on the satellite image input, we decided to center crop our satellite input to 126x126 pixels of the image, reducing training cost at the expense of a reduced spatial context.
    \item Test Time Resizing: Final outputs are 126x126 pixel regions which we upsample to the original 252x252 dimensions using bilinear interpolation for evaluation.
\end{itemize}

We also removed the Water vapor (WV) bands in the dataset, as it was shown by participants of a previous edition of the competition that they contribute little to model performance\cite{2022_park_etal_RainUNetSuperResolutionRain}.
\section{Experiments and Results}
Our methods were evaluated using the Critical Success Index (CSI) on the Weather4cast Stage 2 test set data from 2019-2020. As our focus for this project was developing a competitive model whilst constraining the energy consumption, we limited the number of training epochs to 10.\newline

Our initial experiments were run using the baseline architecture described in Section \ref{model}. We trained one model per region and year, with an estimated median training time of $\sim$2 hours per model. Each model then provided the predicted output for its region and year. This approach yielded a CSI score of 0.05157032, improving the score of 0.04821894 set by the baseline U-Net model at a fraction of its training time and cost.\newline

Given the reduced training epochs of our model, we hypothesised that creating an ensemble of all of the regions could improve the overall predicting performance of our solution. This ensemble model still performed above the original baseline, yet lower than our independent model approach, suggesting that there are some regional features that might be lost by being merged.\newline

In addition to our proposed SmaAt U-Net architecture, we experimented with several variations. These included replacing the CBAM modules with XCAttention\cite{2021_el-nouby_etal_XCiTCrossCovarianceImage} and adding translator blocks\cite{2022_gao_etal_SimVPSimplerBetter} between the encoder and decoder to improve temporal modeling. However, none of these architectural modifications ultimately improved over the performance of our regional approach.\newline
\section{Conclusion}

In this study, we present a modestly sized and computationally efficient model that, despite its compact nature, outperforms baseline models. It is worth noting that our best model, while exhibiting improvement, falls short of achieving substantial gains. However, if put into perspective and considering the number of parameters of a model as a proxy for its computational demand, improving the score of the baseline model at a fraction of its computational cost (our model has $\sim$4.1M parameters vs the $\sim$22M parameters of the baseline model) is an achievement in its own right. If one then considers that this year's top scorers reached a CSI score of 0.1177289, it is only natural to raise the question on how much their score increased relative to the cost of their model.\newline

In summary, our approach underscores the prevailing challenge in balancing model performance and energy efficiency, particularly as the field advances toward larger and more complex architectures. It is essential to acknowledge that, despite achieving advancements in model accuracy, the environmental impact of training these models is often overlooked. We advocate for a holistic approach in evaluating model efficacy, considering not only predictive capabilities but also the ecological footprint, which remains an under-addressed concern in contemporary ML research.

\printbibliography

\end{document}